%% file: acl_latex.tex
\pdfoutput=1

\documentclass[11pt]{article}

\usepackage[final]{acl}

\usepackage{times}
\usepackage{latexsym}
\usepackage{booktabs}
\usepackage[T1]{fontenc}

\usepackage[utf8]{inputenc}
\usepackage{amsmath}
\usepackage{array}
\usepackage{microtype}
\usepackage{multirow}
\usepackage{inconsolata}
\usepackage{enumitem}
\usepackage{amsmath, amssymb}
\usepackage{makecell}

\usepackage{graphicx}
\usepackage[table,xcdraw]{xcolor} 
\usepackage{colortbl}  
\usepackage{booktabs} 
\usepackage{xspace}

\newcommand{\rparagraph}[1]{\vspace{0.0mm}\noindent\textbf{#1.}}

\usepackage[most]{tcolorbox}
\tcbuselibrary{listings, skins}

\newtcolorbox[use counter=lstlisting]{examplebox}[2][]{%
  colback=gray!20,
  colframe=black,
  width=\linewidth,
  boxsep=5pt,
  left=2pt,
  right=2pt,
  top=2pt,
  bottom=2pt,
  fonttitle=\bfseries,
  title={Template~\thelstlisting: #2},
  label={#1},
  enhanced,
  breakable,
  }

\title{MDSEval: A Meta-Evaluation Benchmark for \\Multimodal Dialogue Summarization}

\author{\textbf{Yinhong Liu\textsuperscript{1,2}\thanks{\hspace{1mm} Work done during an internship at AWS AI Labs.},}\quad
    \textbf{Jianfeng He\textsuperscript{1},}\quad
    \textbf{Hang Su\textsuperscript{1},}\quad
    \textbf{Ruixue Lian\textsuperscript{1},}\quad
    \textbf{Yi Nian\textsuperscript{1},}\\
    \textbf{Jake Vincent\textsuperscript{1},}\quad
    \textbf{Srikanth Vishnubhotla\textsuperscript{1},}\quad
    \textbf{Robinson Piramuthu\textsuperscript{1},} \quad
    \textbf{Saab Mansour\textsuperscript{1}} \\
    \textsuperscript{1}AWS AI Labs\\
    \textsuperscript{2}Language Technology Lab, University of Cambridge \\
  yl535@cam.ac.uk, jianfhe@amazon.com, saabm@amazon.es
  }

\begin{document}
\maketitle
\begin{abstract}

Multimodal Dialogue Summarization (MDS) is a critical task with wide-ranging applications. To support the development of effective MDS models, robust automatic evaluation methods are essential for reducing both cost and human effort. However, such methods require a strong meta-evaluation benchmark grounded in human annotations. In this work, we introduce MDSEval, the first meta-evaluation benchmark for MDS, consisting image-sharing dialogues, corresponding summaries, and human judgments across eight well-defined quality aspects. To ensure data quality and richfulness, we propose a novel filtering framework leveraging \textit{Mutually Exclusive Key Information} (MEKI) across modalities. Our work is the first to identify and formalize key evaluation dimensions specific to MDS. We benchmark state-of-the-art modal evaluation methods, revealing their limitations in distinguishing summaries from advanced MLLMs and their susceptibility to various bias.

\end{abstract}

\begin{figure*}
\vspace{-3mm}
    \centering
    \includegraphics[width=\textwidth]{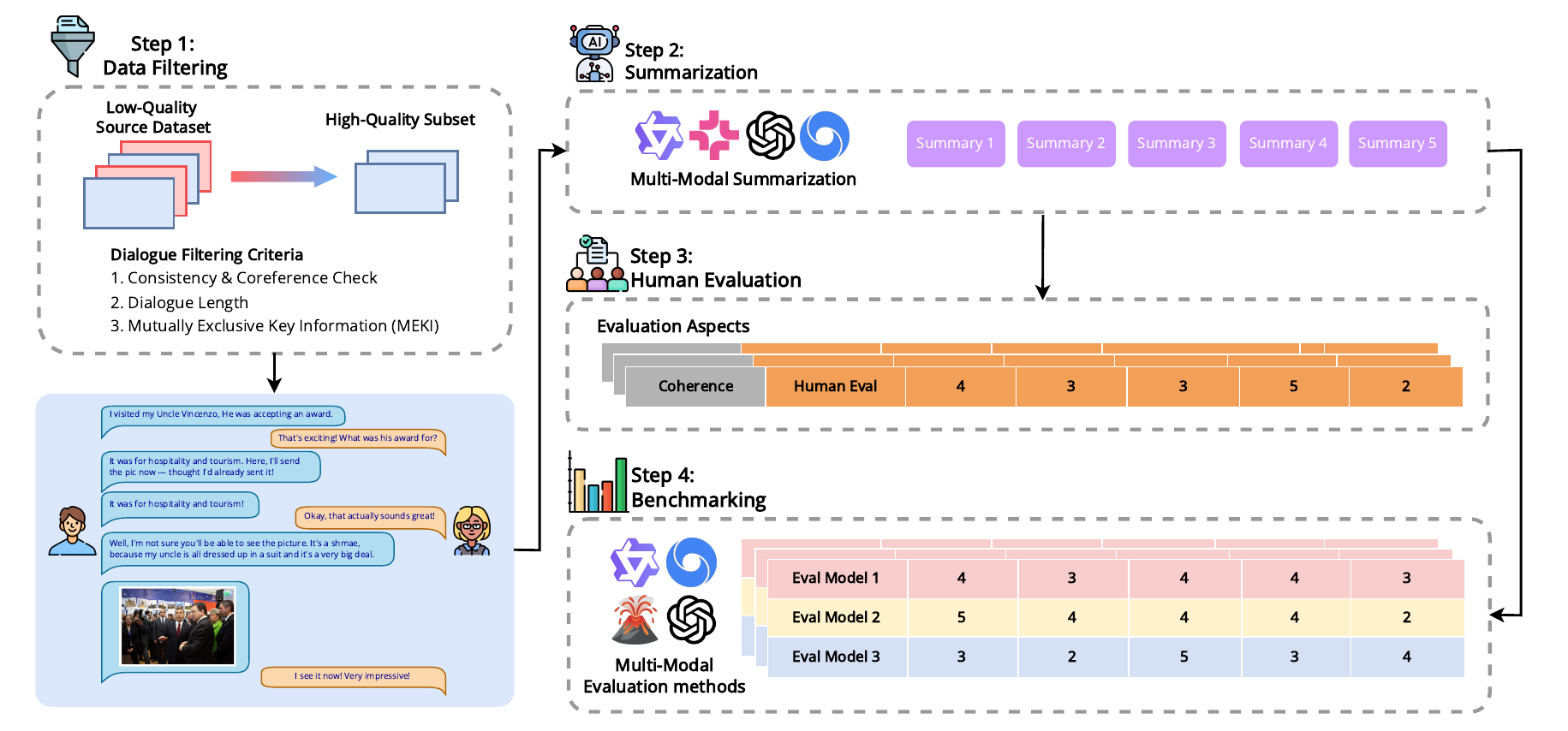}
    \caption{Overview of the MDSEval Curation Pipeline. Step 1: Filter high-quality image-sharing dialogues based on predefined criteria including our proposed MEKI. Step 2: Generate multiple summaries per dialogue using various LLMs and prompting strategies. Step 3: Conduct human evaluations of the generated summaries along key dimensions, including multimodal coherence, content coverage, and faithfulness. Step 4: Benchmark SOTA multimodal LLMs and summarization techniques using our MDSEval dataset.}
    \label{fig:main}
\end{figure*}

\section{Introduction}

Human communication is inherently multimodal, encompassing text, images, videos, and audio. This has led to the emergence of Multimodal Large Language Models (MLLMs), which integrate information across modalities to facilitate more natural and effective human-machine interactions. 
A key application in this space is \textbf{Multimodal Dialogue Summarization} (MDS)~\cite{wang2020vd, kottur2018visual, pan2004mmss}, which aims to distill salient information from multimodal conversations (e.g., Slack chats). 

To accelerate the development of MDS models, reliable automatic evaluation methods are essential, as they enable rapid iteration and significantly reduce the cost of manual assessments. 
However, the development of effective automatic evaluation methods requires a meta-evaluation benchmark—a dataset with detailed human-annotated quality assessments—to serve as ground truth for evaluating the evaluators.

To address this gap, we introduce MDSEval, the first comprehensive meta-evaluation benchmark, specifically for the MDS task.
MDSEval comprises image-sharing dialogues, multiple candidate summaries, and human evaluations across eight quality dimensions. This benchmark enables systematic comparison of evaluation methods, exposes their limitations, and provides actionable insights for developing more accurate and human-aligned assessment techniques for multimodal summarization.

The overall data curation pipeline is illustrated in Figure~\ref{fig:main}.
MDSEval contains 198 high-quality image-sharing dialogues, carefully curated from the PhotoChat \citep{zang2021photochat} and DialogCC \citep{lee2024dialogcc} datasets. To ensure the suitability and challenge level of these dialogues for the summarization task—specifically, that effective summaries must draw on information uniquely provided by both text and images—we introduce a novel data filtering framework based on the\textbf{ Mutually Exclusive Key Information (MEKI)} criterion.
MEKI is designed to identify information that is uniquely conveyed by one modality and not inferable from the other, thereby emphasizing the need for true multimodal comprehension. Empirical analysis shows that MEKI scores correlate strongly with human judgments (Spearman $\rho\approx 0.80$).

Each image-sharing dialogue is paired with five summaries generated by state-of-the-art (SOTA) MLLMs. 
In additional to typical evaluation aspects such as coherence, conciseness and fine-grained faithfulness, we identify and define new aspects focusing on cross-modal  information coverage, information balance and topic progression.
Each summary is annotated by three experienced human experts.
Based on the human annotations, we benchmark latest multimodal evaluation techniques on MDSEval, revealing two critical findings:
\textbf{1)} Current multimodal evaluation methods struggle to differentiate summaries generated by recent LLMs.
\textbf{2)} Existing evaluation methods suffer from significant biases.

MDSEval aims to advance the development of robust, human-aligned multimodal evaluation methods. 
Ultimately, our work paves the way for more sophisticated multimodal conversational agents capable of seamlessly integrating and interpreting information across modalities.
We summarize our main contributions as follows:
\begin{itemize}[noitemsep,topsep=1pt]
    \itemsep 0em
    \item \textbf{MDSEval}: We introduce the first meta-evaluation dataset for MDS, providing human annotations across eight carefully defined evaluation aspects (Section~\ref{sec:human_anno}). We release the benchmark dataset with expert annotations at \url{https://github.com/amazon-science/MDSEval}.
    \item \textbf{MDS Evaluation Aspects}: We define eight novel evaluation aspects specifically tailored for MDS task, with a focus on capturing cross-modal understanding and summary quality.
    \item \textbf{MEKI Criterion}: We propose a novel data filtering framework based onMutually Exclusive Key Information (MEKI). It identifies dialogues where critical information is uniquely conveyed by one modality, and ensures that successful summarization requires genuine multimodal understanding, rather than relying on shortcuts from a single modality.
    \item \textbf{Benchmarking Insights}: Performance of  SOTA multimodal evaluation methods in MDSEval reveals biases and provides insight for advancing assessment.

%
    
\end{itemize}

\section{Background and Related Work}
\rparagraph{Multimodal Dialogue Datasets} 
PhotoChat \citep{zang2021photochat} is an early crowd-sourced dataset simulating human conversations around given images sampled from Open Images V4 \citep{kuznetsova2020open}. MMDialog \citep{feng2023mmdialog}, a large-scale dataset from social media interactions, lacks natural conversational flow due to its fragmented, non-sequential turns \citep{han2023champagne}.
Beyond crowdsourcing, researchers have explored synthesized image-sharing dialogue datasets. MMDD \citep{lee2021constructing} constructs a 45k multimodal dataset by replacing utterances with semantically similar images. DialogCC \citep{lee2024dialogcc} and MAGID \citep{aboutalebi2024magid} employ large language models (LLMs) to detect image-sharing moments in text-only dialogues and retrieve or generate corresponding images. TOAD \citep{liu-etal-2024-toad} proposes a data synthesis framework specifically designed for task-oriented dialogue with multiple speaking styles.

However, a critical and largely overlooked limitation stemming from these construction and synthesis methodologies is the high degree of information overlap between the textual and visual modalities. This redundancy diminishes the datasets' applicability for tasks that require processing complementary information, such as summarizing a dialogue where key details are uniquely present in either the text or the images.

\rparagraph{Multimodal Summarization Dataset} 
Research in multimodal summarization has predominantly focused on news and other fact-intensive domains.
Many datasets are built upon news articles from sources like the DailyMail and CNN, including MSMO \citep{zhu-etal-2018-msmo}, E-DailyMail \citep{chen-zhuge-2018-abstractive}, and the multimodal MM-AVS \citep{fu-etal-2021-mm}.
Other works have drawn from various sources, such as online discussions from Reddit Threads (MREDDITSUM \citep{overbay-etal-2023-mredditsum}), Wikipedia articles with images (REFINESUMM \citep{patil-etal-2024-refinesumm}), or multilingual news reports (M3LS \citep{verma-etal-2023-large}).

However, these resources do not address the specific challenge of summarizing conversational dialogues where images are actively shared between participants. To the best of our knowledge, no prior dataset is designed for this task. The most closely related work, MDS \citep{liu-etal-2024-mds}, incorporates video clips, but its visual modality captures the speakers in a conversational setting rather than images shared as part of the dialogue's content. This scarcity of relevant data is particularly stark when compared to the wide availability of text-only dialogue summarization datasets, such as SAMSum \citep{gliwa2019samsum} and DIALOGSUM \citep{chen2021dialogsum}, highlighting a clear gap in the literature.

\rparagraph{Meta-Evaluation Benchmark} 
A meta-evaluation benchmark assesses the reliability of automatic evaluation metrics by measuring their correlation with human judgments. While meta-benchmarks exist for text-only dialogue summarization, as detailed in Appendix~\ref{app:related_work}, our work contributes the first such benchmark specifically for multimodal image-sharing dialogue summarization.

\section{MDSEval Benchmark}
This section provides a detailed overview of the MDSEval dataset. We describe the multi-stage construction pipeline, illustrated in Figure~\ref{fig:main}, and present the resulting dataset statistics.

\begin{table}[]
    \centering
    \begin{tabular}{l r}
        \toprule
        Statistic & Value \\
        \midrule
        Total number of dialogues & 198 \\
        Summaries per dialogue & 5 \\
        Avg. turns per dialogue & 17.1 \\
        Avg. tokens per dialogue & 209.0 \\
        Evaluation aspects & 8 \\
        Avg. annotators per summary & 2.9 \\
        Avg. sentences per summary & 4.5 \\ 
        \bottomrule
    \end{tabular}
    \caption{Statistics of the MDSEval Dataset.}
    \vspace{-3mm}
    \label{tab:stats}
\end{table}

\subsection{Dataset Statistics}
As shown in Table~\ref{tab:stats}, MDSEval consists of 198 image-sharing dialogues, each paired with five generated summaries. Each summary is evaluated across eight aspects, including fine-grained faithfulness. Most evaluations ($93.1\%$) involved three annotators, with the remaining $6.9\%$ involving two, as explained in Appendix~\S\ref{appen:anno-human}. While nearly all dialogues contain a single shared image, three dialogues include multiple images. Below, we describe the data collection process in detail. All prompts used during the data curation are listed in Appendix~\S\ref{appen:prompt}.

\begin{figure}
    \centering
    \includegraphics[width=\linewidth]{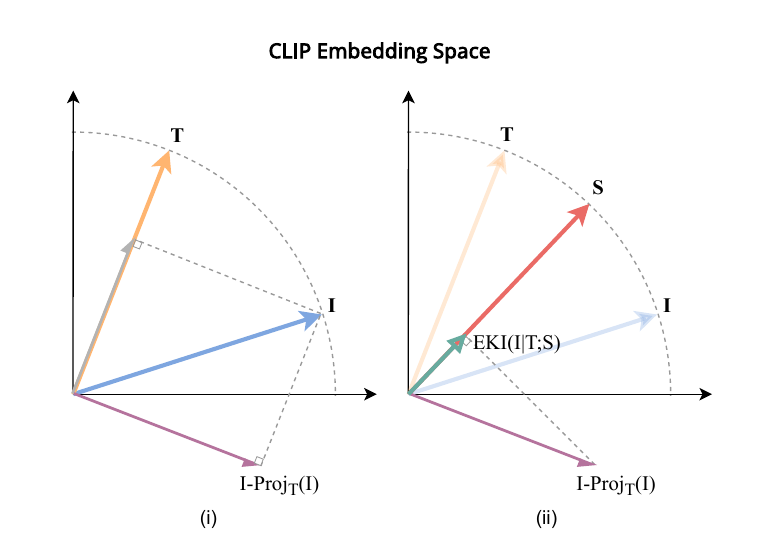}
    \caption{A conceptual illustration of Exclusive Key Information (EKI). Given unit-normalized CLIP embeddings for a text ($T$), image ($I$), and pseudo-summary ($S$), we first compute the Exclusive Information (EI) of the image as the component orthogonal to the text:$\operatorname{EI}(I|T) = I - \operatorname{Proj}_T(I)$. The EKI is then calculated by projecting this resulting EI onto the pseudo-summary embedding $S$.}
    \label{fig:MEKI}
\end{figure}

\begin{figure*}[t!]
\centering
\begin{minipage}{0.48\textwidth}
    \vspace{-4mm}
        \centering
        \includegraphics[width=\textwidth]{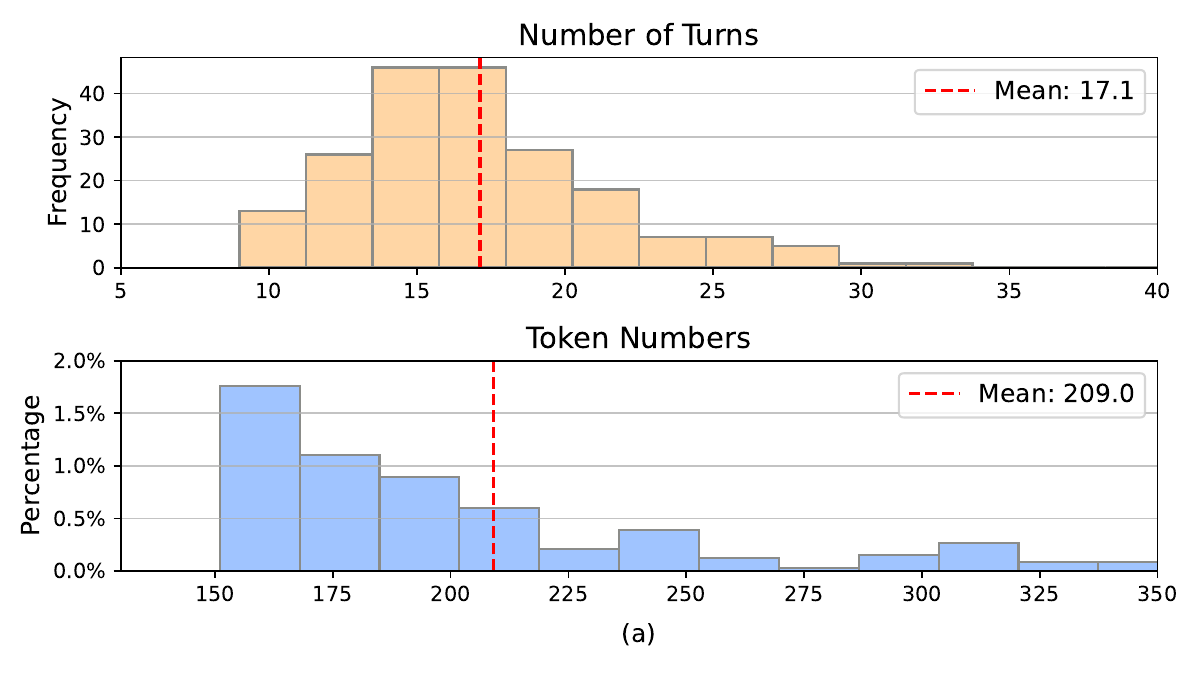}
\end{minipage}%
\begin{minipage}{0.55\textwidth}
    \vspace{-3mm}
        \centering
        \includegraphics[width=\textwidth]{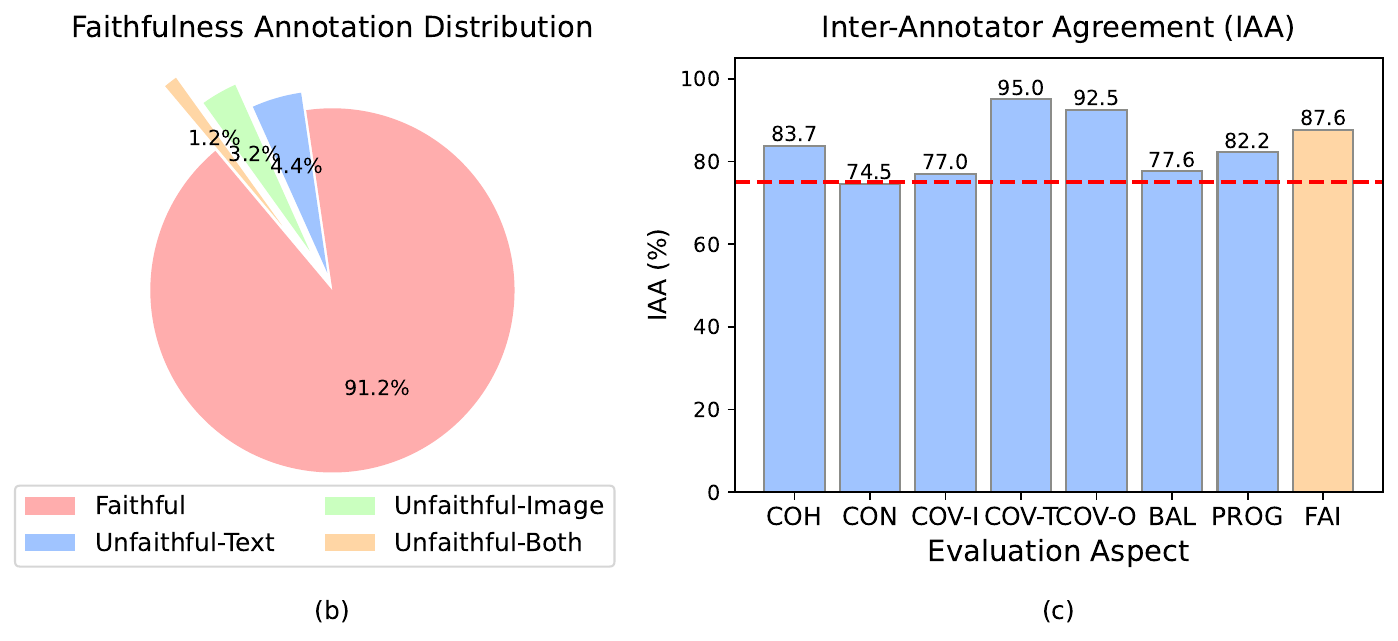}
\end{minipage}%
\vspace{-3mm}
\caption{Statistics of MDSEval: (a) Upper: Distribution of the number of dialogue turns. Lower: Distribution of the number of dialogue tokens. (b) Faithfulness distribution at the summary level. (c) Inter-annotator agreement (IAA) for all evaluation aspects (see Section~\ref{sec:human_anno} for the details about evaluation aspects). All aspects show strong inter-annotator agreement, with adjacent agreement rates exceeding $74.5\%$.}

\label{fig:dist}

\end{figure*}

\subsection{Step 1: Data Filtering Framework}

We begin with two existing image-sharing dialogue datasets: PhotoChat \citep{zang2021photochat}, which consists of ~12K dialogues simulated by two parties chatting about an image, and DialogCC \citep{lee2024dialogcc}, which contains ~45K social dialogues where certain utterances are replaced with semantically similar images. However, both datasets include noisy dialogues, and not all are suitable or sufficiently challenging for summarization due to the overlapping information between modalities. To address this, we propose a filtering framework to extract a high-quality subset of multimodal dialogues from these two datasets for our MDSEval dataset, based on two key criteria.

\rparagraph{Suitability} A multimodal dialogue is considered suitable for summarization task if it 1) contains sufficient information, which we ensure by removing dialogues with fewer than 150 tokens, and 2) maintains consistency between shared images and text. Due to inconsistencies in the source datasets, we employ Claude-3.5-Sonnet \citep{claude2024} for an initial automatic check to assess image-dialogue coherence and cross-modal coreferences. The prompts used are listed in Appendix~~\S\ref{appen:prompt}. After filtering, around 3K dialogues remain from the original 57K.

\rparagraph{MEKI Metric} 
To ensure the multimodal dialogues are sufficiently challenging for summarization, they should contain key information uniquely conveyed by each single modality — meaning it cannot be inferred from the other. To quantify this, we introduce Mutually Exclusive Key Information (MEKI) as a selection metric.

As shown in the conceptual demonstration in Figure~\ref{fig:MEKI}, we embed both the image and textual dialogue into a shared semantic space using the CLIP model\footnote{CLIP ViT-H-14-378-quickgelu from OpenCLIP \citep{ilharco_gabriel_2021_5143773}}, denoted as vectors  $I\in \mathbb{R}^N$ and $T \in \mathbb{R}^N$. $N$ is the embedding dimension. Since CLIP embeddings are unit-normalized, we maintain this normalization for consistency.
To measure \textit{Exclusive Information} (EI) in $I$ that is not present in $T$, we compute the orthogonal component of $I$ relative to $T$:
\begin{equation}
   I_T^\perp = I - \operatorname{Proj}_T(I) = I -  \frac{\langle I, T\rangle}{\langle T, T\rangle} T,
\end{equation}
where $\langle \cdot , \cdot \rangle$ denote the dot product.

Next, to identify \textit{Exclusive Key Information} (EKI) — crucial content uniquely conveyed by one modality — we generate a pseudo-summary $S$ using Claude-3.5-Sonnet, which extracts essential dialogue and image details. This serves as a reference proxy rather than a precise summary, helping distinguish key information. We embed and normalize $S$ in the CLIP space and compute:
\begin{equation}
  \operatorname{EKI}(I|T; S) =  
  \left\| \frac{\langle I_T^\perp, S\rangle}{\langle S, S\rangle} S \right\|,
\end{equation}
which quantifies the extent of exclusive image-based key information. Similarly, we compute $\operatorname{EKI}(T|I; S)$ for textual exclusivity.
Finally, the MEKI score aggregates both components:
\begin{align}
    \nonumber
    \operatorname{MEKI}(I, T; S) &= \lambda \operatorname{EKI}(I|T; S) \\ 
                                 &+ (1-\lambda)\operatorname{EKI}(T|I; S),
\end{align}
where $\lambda=0.3$, chosen to balance the typically higher magnitude of the exclusivity term in text-based information, ensuring that the average magnitudes of both terms are approximately equal. 
We also implement and explore several variants of the MEKI criterion, validating and comparing their effectiveness through human judgment studies (details in Appendix~\S\ref{appen:meki}. 
We rank the filtered $3K$ dialogues by MEKI scores and select the top $300$ as the most challenging dialogues for later processing.

\subsection{Stage 2: Summary Generation}

In this stage, we generate summary candidates for each dialogue to evaluate the effectiveness of SOTA evaluation methods. To ensure a diverse quality spectrum, we use four modern MLLMs—allenai/Molmo-72B-0924 \citep{deitke2024molmo}, GPT-4o-mini \citep{openai2024gpt4omini}, Gemini-1.5-flash \citep{team2024gemini}, and Qwen-vl-max \citep{bai2023qwen}, along with three prompting strategies: zero-shot prompting (direct generation without examples), In-Context Learning (ICL) \citep{dong2022survey} (with a reference summary demonstrating the desired summary style), and guidance prompting (with detailed instructions specifying the summary standard). 
The prompt templates are detailed in Appendix~\S\ref{appen:prompt}. 

As a result, a total of 12 \textit{generation setups} emerge from combining different LLMs and prompting techniques. However, due to annotation capacity, we select only five summary candidates per dialogue.
To maximize diversity, we choose five setups that yield the highest total pairwise distance among their generated summaries. Concretely, let $c\in C$ denote a combination, which combines 5 generation setups. The $C$ represents the set of all possible combinations. Since we have 12 LLM-prompt setups in total, the number of all possible combinations is $\lvert C \rvert = \binom{12}{5}=792$.
The optimal setup combination, $c^*$, is determined by:
\begin{align}
\nonumber    c^* & = arg\max_{c\in C} D(c) = arg\max_{c\in C} \sum_{1\leq i<j\leq \lvert c \rvert} d_{ij}
\end{align}
where $d_{ij}$ represents the Euclidean distance between the embedding vectors\footnote{OpenAI's text embedding model, text-embedding-3-large} of summaries $i$ and $j$. 
Table~\ref{tab:setup} shows the final selection of setup combination, $c^*$, which maximize total pairwise distance.

\begin{table}
    \centering
    \begin{tabular}{cll}
    \hline
        Setup & Models & Prompting\\
    \hline
       1  & allenai/Molmo-72B-0924 & guidance\\
       2  & allenai/Molmo-72B-0924 & ICL\\
       3  & qwen-vl-max & ICL\\
       4  & gpt-4o-mini & guidance\\
       5  & gemini-1.5-flash & zeroshot\\
   \hline
    \end{tabular}
    \caption{The selected combination of generation setups for summary generation, which produces summaries with the highest total pairwise distance.}
    \label{tab:setup}
\end{table}

\input{tables/table_pointwise}

\subsection{Stage 3: Human Annotations} \label{sec:human_anno}
To evaluate the quality of summaries, we define evaluation aspects tailored to the Multimodal Dialogue Summarization task. These aspects are designed to address the multimodal nature of the task and identify areas where state-of-the-art MLLMs often struggle. The evaluation aspects are as follows:

\noindent\textbf{(1) Multimodal Coherence (COH)} measures how naturally the summary integrates and organizes information from both visual elements of the image and the textual content of the dialogue.

\noindent\textbf{(2) Conciseness (CON)} assesses how efficiently the summary conveys essential information without unnecessary verbosity or redundancy.

\noindent\textbf{(3-5) Multimodal Coverage (COV)} evaluates the extent to which the summary captures the breadth and depth of key information from the source material. Key information includes details that describe the core event or significantly contribute to the main ideas of the dialogue. Coverage is further divided into three sub-aspects based on modalities:
Visual Critical Information Coverage (COV-I), Textual Critical Information Coverage (COV-T), and Overall Critical Information Coverage (COV-O).

\noindent\textbf{(6) Multimodal Information Balancing (BAL)} evaluates how well the summary balances information from different modalities. Overemphasis on visual information may result in an image caption, while excessive focus on textual information may lead to a pure textual summary.

\noindent\textbf{(7) Topic Progression (PROG)} evaluates the ability of the summary to accurately capture the flow of topics discussed in the dialogue. A well-structured summary should ensure smooth transitions between topics and correctly associate shared images with their corresponding parts of the dialogue.

\noindent\textbf{(8) Multimodal Faithfulness (FAI)} is annotated at the \textit{sentence level}. Each summary sentence is evaluated on whether it accurately reflects the content of the original dialogue and associated images without introducing incorrect, fabricated, or misleading information. This includes factual details such as: relationships between entities, descriptions of objects, spatial relations and other key elements present in the dialogue or images. 
The faithfulness annotation labels include: \textit{Faithful}, \textit{Not faithful to text}, \textit{Not faithful to image}, and \textit{Not faithful to both}. By aggregating the sentence-level faithfulness annotations as described in Appendix~\S\ref{appen:faith_aggre}, we also derive summary-level faithfulness annotations.

All aspects, except for faithfulness and information balancing, are rated on a 5-point Likert scale. Information balancing is assessed on a 1-7 bipolar scale, where 1 indicates an overemphasis on textual information and 7 on visual information. Faithfulness is classified into four categories introduced above. 
Detailed annotation guidelines and the annotation interface are provided in Appendix~\S\ref{appen:anno}. 
The Inter-Annotator Agreement (IAA) is presented in Figure~\ref{fig:dist}c, utilizing the adjacent agreement rate, which measures the percentage of score pairs that differ by no more than 1. For the faithfulness assessment, IAA is determined based on the exact agreement rate across all annotators. In instances where a majority consensus is not reached for faithfulness, an additional round of annotation is carried out, as described in Appendix~\S\ref{appen:faith_anno}. Overall, we observe that a strong level of agreement is achieved, with all IAA values exceeding $75\%$.

\subsection{Stage 4: Benchmarking}
We evaluate three state-of-the-art multimodal assessment methods on MDSEval: MLLM-as-a-Judge \citep{chenmllm} uses a MLLM to assign scores or make pairwise comparisons in a zero-shot manner. Image-to-Prompt \citep{guo2023images} generates detailed image descriptions via an MLLM, then pairs them with the textual query for a text-only LLM to render the final judgment. LLaVA-Critic \citep{xiong2024llava} is a supervised fine-tuned version of LLaVA \citep{liu2024visual}, designed specifically for multimodal evaluation.

Additionally, we introduce Checklist-CoT, a multimodal checklist-based Chain-of-Thought (CoT) prompting baseline method. It guides MLLMs through explicit structured discussion: first discussing key concepts in each modality via a checklist, then integrating cross-modal insights before making a final evaluation.

Among these, MLLM-as-a-Judge, Image-to-Prompt, and Checklist-CoT are training-free frameworks adaptable to various base LLMs. We benchmark them using three leading MLLMs: GPT-4o-mini \citep{openai2024gpt4omini}, Gemini-1.5-flash \citep{team2024gemini}, and Qwen-vl-max \citep{bai2023qwen}. Full prompt details are provided in Appendix~\S\ref{appen:prompt}.

\input{tables/table_pairwise}

\section{Benchmarking Results}\label{sec:benchmark}
\subsection{Setup and Metrics}
\rparagraph{Evaluation Paradigms}
Following \citet{zheng2023judging}, we adopt two common evaluation paradigms: score-based pointwise evaluation and pairwise comparison. As described in Section~\ref{sec:human_anno}, coherence, conciseness, coverage, and topic progression are rated on a Likert scale (1–5), multimodal information balance on a 7-point bipolar scale, and faithfulness using a 4-way classification. 
We also perform pairwise evaluations for all aspects except faithfulness, which should be assessed independently without comparison. The corresponding evaluation prompts are provided in Appendix~\S\ref{appen:prompt}.

\rparagraph{Human Preferences} 
To aggregate annotations across multiple human evaluators, we compute the average score in score-based evaluations. Faithfulness labels are determined by majority vote. For pairwise comparisons, the summary with the higher averaged score is considered preferred.

\rparagraph{Metrics}
For score-based evaluation, we use Spearman’s rank correlation coefficient ($\rho$) to measure alignment with human judgments, computed per dialogue and averaged across instances \citep{liu2023g, zhong2022towards}. We also report Mean Squared Error (MSE) to assess prediction robustness. For pairwise evaluation, accuracy (Acc.) serves as the primary metric. Given the skewed label distribution in faithfulness evaluation, we use Balanced Accuracy (BAcc.) and F1-score (F1) to account for class imbalance.

\subsection{Main Results}
\begin{figure*}[h]
\vspace{-3mm}
    \centering
    \includegraphics[width=\linewidth]{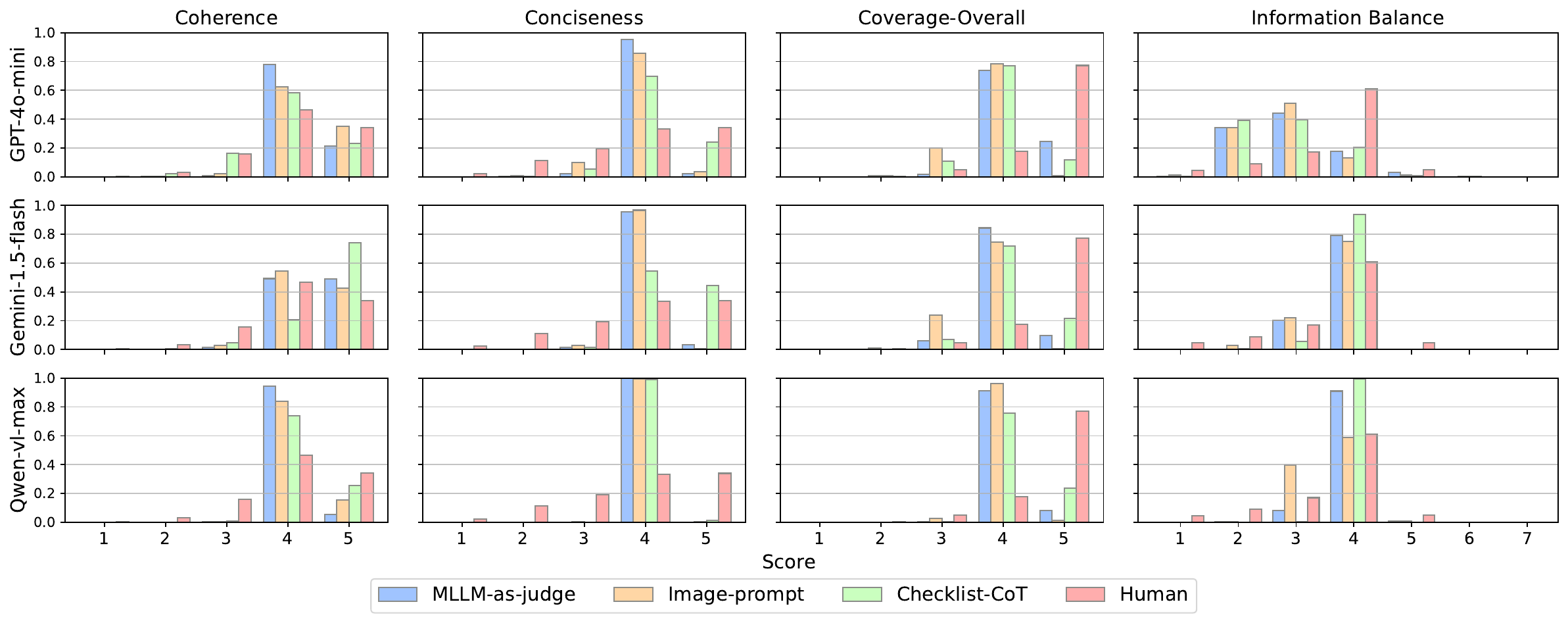}
    \vspace{-4mm}
    \caption{\textit{Multimodal evaluation methods exhibit significant score distribution bias}, with most evaluations concentrated on score of $4$. This figure presents the score distributions of evaluation methods compared to human score distributions across four selected evaluation aspects, demonstrating a strong misalignment with human assessments.}
    \label{fig:score_bias}
\end{figure*}

\rparagraph{MLLM-based evaluators fail to align with human judgments of summary quality}
Our results in Tables~\ref{tab:pointwise} and \ref{tab:pairwise} show that these methods have a consistently weak correlation with human preferences, 
We identify a primary cause for this misalignment: a systematic bias towards score concentration. As visualized in Figure~\ref{fig:score_bias}, evaluators tend to "hedge" their assessments, producing scores within a very limited range. This lack of variance severely limits their ability to discriminate between summaries, rendering them ineffective for evaluating the nuanced differences in quality among outputs from advanced LLMs. Furthermore, the low and unstable Spearman correlation values suggest that rank-based metrics are particularly unreliable here. In contrast, we find that MSE offers a relatively more stable measure of evaluator performance under these conditions.

Image-prompting is particularly ineffective for assessing visual information coverage. We posit that this shortcoming stems from information loss incurred when translating images into textual descriptions for the MLLM. These errors subsequently propagate through the evaluation, compromising the final judgment.
Additionally, Table~\ref{tab:pointwise} indicates that GPT-4o-mini struggles to interpret the bipolar scale used for the Information Balancing criterion, a systematic bias we further analyze in Section~\ref{sec:bias}. We also observe that CoT prompting does not yield consistent improvements. For instance, while the checklist-CoT method performs well on Coverage, it is the worst-performing variant for Coherence.

For pairwise evaluations, visual information coverage is assessed more reliably than other aspects. We hypothesize that this is due to MLLMs being primarily trained on image captioning and Visual Question Answering (VQA) tasks, which focus on image content discussion. In contrast, other aspects of evaluation require interpreting visual data as contextual cues rather than the primary focus. As a result, MLLMs struggle with instructions that deviate significantly from their training paradigm.

\begin{figure*}
    \centering
    \includegraphics[width=\linewidth]{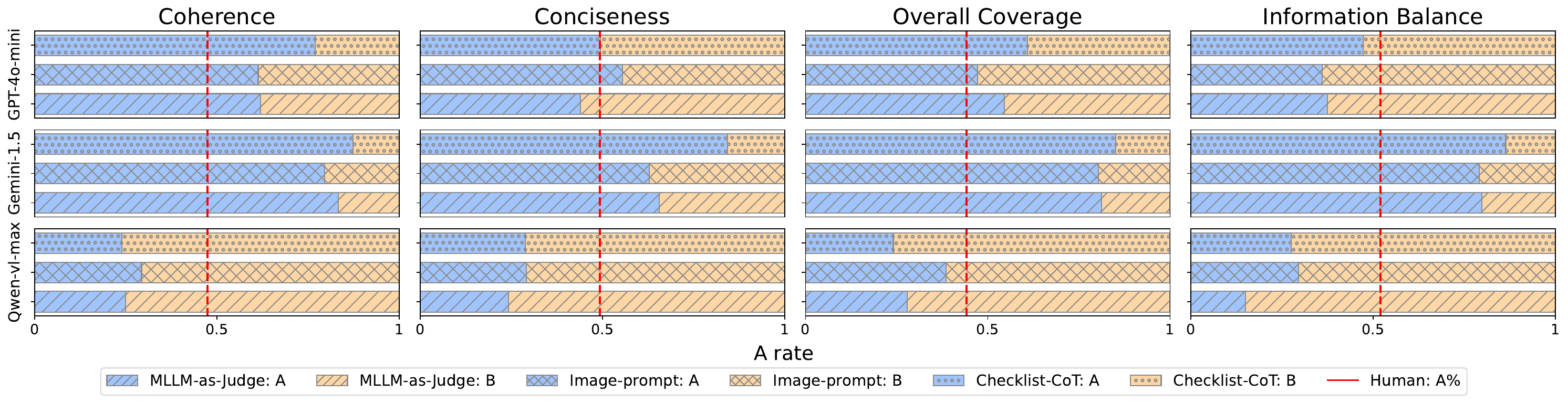}
    \vspace{-4mm}
    \caption{\textit{Different base MLLMs exhibit varying degrees of positional bias.} This figure shows the overall preference ratio between options A and B in pairwise comparisons. Compared to the preference ratios from human annotations, GPT-4o-mini demonstrates the least bias, while the other two MLLMs show stronger biases. Gemini-1.5-flash favors option A, whereas Qwen-vl-max favors option B.}
    \label{fig:pairwise_bias}
    \vspace{-3mm}
\end{figure*}

\rparagraph{Comparison of Different Base MLLMs}
Among the tested models, Qwen-VL-Max exhibits slightly lower error in score-based evaluations in most aspects, while GPT-4o-mini performs marginally better in pairwise comparisons. However, these differences are minimal. The LLaVA-Critic model performs comparably to GPT-4o-mini, likely because it was trained on annotations from GPT-4o, which serves as its upper performance bound.
Overall, our findings suggest that current MLLMs fail to provide human-aligned judgments when evaluating text generated by advanced LLMs.

\subsection{Faithfulness Benchmark}
We evaluate faithfulness at two granularities: \textit{summary level} and fine-grained \textit{sentence level}, as presented in Table~\ref{tab:faithfulness}. 
Overall, faithfulness evaluation performance remains low, with all balanced accuracy scores falling below $30\%$. This challenge primarily stems from the inherent complexity of inference in multimodal contexts, as well as the imbalanced distribution of faithfulness labels (Figure~\ref{fig:dist}b).
Notably, summary-level evaluations outperform sentence-level assessments. We hypothesize that this is due to the lack of previous summary context at the sentence level, requiring MLLMs to rely more on deductive reasoning — an inherently more difficult task.

\input{tables/table_faithful}

\subsection{Performance Analysis: Evaluation Bias} \label{sec:bias}
Inductive biases, such as positional bias \citep{wang-etal-2024-large-language-models-fair, zhou-etal-2024-fairer, liu2024aligning2}, selection bias \citep{zheng2024large}, and score distribution bias \citep{liu2024aligning}, have been widely studied in text-only LLM-based evaluation. However, these biases remain largely underexplored in the context of MLLM evaluation. In this section, we analyze the impact of these biases when evaluating models using MDSEval.

Figure~\ref{fig:score_bias} illustrates the bias in score distribution in four selected aspects, compared to human score distributions. Regardless of the evaluation method applied, all MLLMs exhibit strong biases in score distribution. Notably, for most aspects, the predicted scores are disproportionately concentrated at 4, deviating significantly from human annotations. This finding reaffirms that current MLLM evaluation methods struggle to effectively assess summaries generated by state-of-the-art models. A particularly interesting case arises when using GPT-4o-mini to evaluate the bipolar cross-modality information balance aspect—the predicted score distribution is noticeably skewed toward text-heavy outputs, which aligns with the high mean squared error (MSE) observed in Table~\ref{tab:pointwise}.

Figure~\ref{fig:pairwise_bias} shows positional bias in pairwise comparisons, contrasting model preferences against human judgments.
The red dashed lines indicate the human preference ratio derived from annotations. Note that for clarity, each pairwise comparison is represented only once without switching the option order.
As shown, GPT-4o-mini exhibits the least positional bias, while Gemini-1.5-flash always favors the first option, and Qwen-vl-max demonstrates a preference for the second option. Notably, different evaluation methods do not mitigate these positional biases, suggesting that systematic biases persist across evaluation frameworks.

\section{Conclusion}
In this work, we introduce MDSEval, the first meta-evaluation benchmark for multimodal dialogue summarization. MDSEval includes human annotations across eight carefully defined evaluation aspects. We detail our data curation pipeline and propose novel metrics and considerations—such as the Mutually Exclusive Key Information (MEKI) scores—which will inspire future research. Furthermore, we assess the performance of several SOTA multimodal evaluation methods on MDSEval and find that current MLLMs still struggle to deliver human-aligned judgments when evaluating summaries generated by the latest MLLMs.

\section{Ethical Consideration} 
This work provides a human-annotated benchmark for multimodal dialogue summarization using publicly available datasets, which minimizes immediate ethical concerns related to data privacy and accessibility. However, future extensions of this research into domains containing personal or sensitive information could pose privacy risks if not carefully managed. We strongly recommend that any such expansions implement stringent ethical guidelines, including robust anonymization techniques and informed consent protocols.

Regarding the licenses, the datasets integral to our work are utilized in adherence to their respective licenses \citep{zang2021photochat,lee2024dialogcc}.

\section{Limitations}
One limitation of MDSEval is its focus solely on chitchat-style dialogues. While this serves as a useful starting point, it does not capture the full diversity of real-world dialogue scenarios. Expanding to more practical domains—such as commercial customer service or formal workplace conversations—could enhance the benchmark's applicability and relevance.
Additionally, MDSEval currently supports only text and image modalities. Incorporating richer modalities such as video and audio, which frequently appear in real-world multimodal dialogues, would make the benchmark more realistic and comprehensive.

\bibliography{custom}

\clearpage
\appendix

\section{Additional Related Works}
\label{app:related_work}
\noindent\textbf{Meta-Evaluation Benchmark.} Meta-evaluation benchmarks assess automated metrics by comparing metric outputs (e.g., rating) to human judgments across multiple responses. These benchmarks examine the alignment with human evaluations, particularly in tasks with diverse valid outputs such as machine translation and summarization. These benchmarks, incorporating coherence, relevance, fluency, and factual consistency, underscore the limitations of traditional automated metrics from the human annotation perspective.

In machine translation, the WMT Metrics Shared Task \citep{bojar2017findings} exposed the limitations of BLEU, which relies on word overlap. For summarization, SummEval \citep{fabbri2021summeval} revealed that ROUGE often fails to capture summary quality. RealSumm \citep{deutsch2021towards} and QAEval \citep{deutsch2021towards} focus on factual consistency, assessing a summary’s alignment with its source using human-annotated accuracy scores.
In dialogue evaluation, USR \citep{mehri2020usr} proposed a reference-free framework leveraging semantic similarity and dialogue context, outperforming traditional metrics like BLEU. mLLM-Eval \citep{zhuang2024automatic} extends meta-evaluation to multimodal summarization by benchmarking metrics on the MSMO task with human-annotated scores.
In the multimodal dialogue summarization, to the best of our knowledge, we are the first to contribute a meta-benchmark.

\section{MEKI Evaluation}
\label{appen:meki}

In this section, we present our implementation and evaluation of the MEKI score, which aims to measure the alignment of Key Exclusive Information (EKI) between images and corresponding dialogues. We explore four distinct implementations of MEKI and assess their effectiveness by comparing their outputs to human and large language model (LLM) evaluations.

\begin{itemize}[noitemsep,topsep=1pt]
    \itemsep 0em
    \item \textbf{Implementation 1:} MEKI is computed directly using the CLIP embeddings of the paired image and its corresponding dialogue text. This approach leverages CLIP’s cross-modal embedding space to measure the semantic alignment between the visual and textual content without any additional processing.

    \item \textbf{Implementation 2:} Before computing MEKI, we summarize the dialogue into a concise list of core statements, filtering out non-informative or redundant expressions. We then use CLIP to compute the semantic similarity between the image and the resulting dialogue summary. This method aims to reduce noise from casual or filler dialogue that may obscure key information.

    \item \textbf{Implementation 3:} We extract content statements from both the image and the dialogue using separate summarization processes. The image is described in textual form (e.g., via captioning), and the dialogue is summarized into discrete informative statements. We then compute MEKI using a textual embedding model (e.g., OpenAi's text embedding model, text-embedding-3-large) to measure similarity between the two sets of statements. This implementation removes dependency on joint image-text embeddings and focuses instead on textual alignment.
    
    \item \textbf{Implementation 4:} This implementation performs fine-grained content comparison through logical entailment. Both the image and the dialogue are first converted into structured lists of content statements. For EKI-image, we examine each image-derived statement to determine whether it is entailed by the dialogue statements using a large language model. The percentage of statements not entailed by the dialogue is taken as the EKI-image score. Similarly, for EKI-dialogue, we assess which dialogue statements are not entailed by the image-derived content. This method reflects an information disentanglement approach commonly used in prior multimodal related tasks \citep{sanders-etal-2024-tv, suzuki-etal-2019-multimodal, liu-etal-2023-interpretable}.
\end{itemize}

To determine which MEKI implementation most accurately captures EKI, we conduct verification using both human annotators and an LLM. Specifically, we evaluate how well each MEKI variant correlates with human and LLM judgments of EKI for both image and dialogue components within paired data. Annotators are instructed to rate the EKI independently for the image and the dialogue on a scale from 1 (very low exclusive information) to 5 (very high exclusive information). These evaluations are performed on a set of 20 randomly selected dialogue-image pairs.
We report the Spearman correlations in Table~\ref{tab:eki_comparison}.

For human evaluation, each dialogue is evaluated by three independent annotators. To assess the consistency of the annotations, we compute inter-annotator agreement using Krippendorff’s Alpha. The resulting agreement scores are 0.73 for EKI-image and 0.69 for EKI-dialogue, indicating a substantial level of reliability in the human evaluations.

\begin{table}[h]
    \centering
    \resizebox{\linewidth}{!}{%

        \begin{tabular}{ lcccc }
            \hline
            \multirow{2}{*}{MEKI Variant} & \multicolumn{2}{c}{Human} & \multicolumn{2}{c}{LLM} \\
                                             & EKI-I & EKI-D  & EKI-I & EKI-D \\
            \cmidrule(lr){1-1} \cmidrule(lr){2-3} \cmidrule(lr){4-5}
            Variant 1 & \textbf{0.81} & \textbf{0.76} & 0.69 & 0.74 \\ 
            Variant 2 & 0.78 & 0.66 & \textbf{0.71} & 0.75 \\ 
            Variant 3 & 0.72 & 0.73 & 0.65 & \textbf{0.78} \\ 
            Variant 4 & 0.76 & 0.65 & 0.68 & 0.72 \\ 
            \hline
        \end{tabular}
    }
    \caption{Correlation of MEKI Implementations with Human and LLM Judgments on EKI}
    \label{tab:eki_comparison}
\end{table}

\section{Prompt Templates}\label{appen:prompt}

\begin{examplebox}{Multimodal Summarization - Zeroshot}
You are tasked with summarizing a dialogue that includes shared images. Your goal is to create a smooth, coherent summary that integrates both the textual conversation and the visual elements shared during the dialogue.
First, you will be presented with the textual dialogue:\\

<dialogue>\{\{ dialogue \}\}</dialogue>\\

Please provide your summary within <summary> tags. Remember to create a smooth, integrated summary that effectively combines both the textual and visual elements of the conversation.\\

<summary> [Your summary goes here] </summary>
\end{examplebox}

\begin{examplebox}{Multimodal Summarization - Guidance}
You are tasked with summarizing a dialogue that includes shared images. Your goal is to create a smooth, coherent summary that integrates both the textual conversation and the visual elements shared during the dialogue.\\
First, you will be presented with the textual dialogue:\\

<dialogue>\{\{ dialogue \}\}</dialogue>\\

To create an effective summary, follow these steps:\\
1. Carefully read through the entire dialogue, paying attention to the main topics discussed and any key points or decisions made.\\
2. Take note of where images are mentioned or shared in the conversation and how they relate to the discussion.\\
3. Analyze the image descriptions and consider how they contribute to or illustrate the topics being discussed.\\
4. Identify the main themes or subjects of the conversation, including how the images support or enhance these themes.\\
5. Create a concise summary that: Captures the essence of the conversation and omit the unimportant details. Integrates references to the shared images naturally. c. Maintains a logical flow of ideas. Highlights any important conclusions or outcomes\\
6. Ensure that your summary is coherent and reads smoothly, transitioning naturally between textual content and image references.\\
7. Keep the summary concise, aiming for about 3-4 sentences, unless the complexity of the dialogue requires a slightly longer summary.\\
Please provide your summary within <summary> tags. Remember to create a smooth, integrated summary that effectively combines both the textual and visual elements of the conversation.\\

<summary> [Your summary goes here] </summary>
\end{examplebox}

\begin{examplebox}{Multimodal Summarization - ICL}
You are tasked with summarizing a dialogue that includes shared images. Your goal is to create a smooth, coherent summary that integrates both the textual conversation and the visual elements shared during the dialogue.\\

Below we show an example of a coherent and natural dialog summary:\\
<example>\\
Conversation: Two friends, Sarah and John, discuss their weekend plans. Sarah mentions wanting to go hiking at a nearby trail, while John prefers to visit a new art exhibit in the city. They consider combining the two activities by hiking in the morning and visiting the exhibit in the afternoon. Sarah is unsure if they'll have enough time, but John reassures her that the timing should work. They decide to meet at 8 AM for the hike and check out the exhibit afterward if they’re not too tired.
Summary: Sarah and John plan their weekend activities, agreeing to go hiking in the morning and potentially visit an art exhibit afterward. They coordinate meeting at 8 AM and decide to play it by ear based on how they feel after the hike.\\
</example>\\

Now, you will be presented with the textual dialogue:\\
<dialogue> \{\{ dialogue \}\} </dialogue>\\

Please provide your summary within <summary> tags. Remember to create a smooth, integrated summary that effectively combines both the textual and visual elements of the conversation.\\

<summary> [Your summary goes here] </summary>
\end{examplebox}

\begin{examplebox}{Multimodal Pseudo-Summarization}
You are tasked with summarizing a multiturn dialogue that includes a shared image. The dialogue and a detailed description of the image will be provided. Your goal is to create a concise and coherent summary that captures the essential information from the image-sharing conversation.\\

First, you will be provided with the full text of the dialogue:\\
<dialogue> \{\{dialogue\}\} </dialogue>\\
                
Next, you will be given a detailed description of the image that was shared during the dialogue:\\
<image\_description> \{\{Image descriptions\}\} </image\_description>\\

To create an effective summary, please follow these guidelines:\\
1. Focus on the essential information related to the image-sharing conversation.\\
2. Include relevant details about the image only if they are important information and should be included in the dialogue summary.\\
3. Remember that the image may not always be the central topic of the conversation, but could be part of it.\\
4. Keep the summary concise, aiming for no more than 3-4 sentences.\\
5. Ensure the summary is coherent and flows logically. There is no need to explicitly mention the photo sharing action.\\

After analyzing the dialogue and image description, compose your summary. Present your summary within <summary> tags. Remember to focus only on the essential aspects of the image-sharing conversation, creating a concise and coherent overview of the interaction.\\

<summary> [Your summary goes here] </summary>
\end{examplebox}

\begin{examplebox}{Dialogue Cross-Modality Consistency Check}
You will be given a multiturn dialog and descriptions of photos shared during the dialog. Your task is to examine if the images are appropriate for the corresponding dialog.\\

Here is the dialog:\\
<dialogue> \{\{dialogue\}\} </dialogue>\\

Here are the descriptions of the shared photos:\\
<image\_description> \{\{Image descriptions\}\} </image\_description>\\

Your goal is to identify any potential inconsistencies or irrelevance between what is discussed in the dialog and what is shown in the images. 
The consistency is low when the images are not suitable for the dialog. 
For example, if the dialog mentions a pregnant mother but the image shows parents with a child, this would be a conflict as the baby has not been born yet in the dialog.\\

You should assign a score between 1 and 5, where:\\
1 = Serious conflict or completely irrelevant\\
2 = Significant conflict\\
3 = Moderate conflict or relevance\\
4 = Minor conflict \\
5 = No conflict or very relevant\\

Before providing your final answer, use the scratchpad to analyze the dialog and image descriptions, noting any potential conflicts or inconsistencies.\\
<scratchpad> [Your analysis here] </scratchpad>\\

After your analysis, provide your final score inside <score> tags.\\
<score> [Your score here] </score>
\end{examplebox}

\begin{examplebox}{Dialogue Cross-Modality Coreference Check}
You will be given a multiturn dialog. Your task is to determine if there is cross-modality coreference in the textual dialog that refers to people in a shared photo (note that the photo itself is not provided).\\

Cross-modality coreference occurs when expressions in the text refer to entities that are present in another modality (in this case, a photo).\\

Here is the dialog:\\
<dialogue> \{\{dialogue\}\} </dialogue>\\

Carefully analyze the dialog for any instances where pronouns or other referring expressions might be pointing to people in a photo that the conversational participants can see but isn't explicitly described in the text.\\

Provide your reasoning within <reasoning> tags. Consider the following:\\
1. Are there any pronouns or referring expressions that don't have a clear antecedent in the text?\\
2. Does the conversation imply the presence of a shared visual context (like a photo)?\\
3. Are there any statements or questions that seem to refer to visual information not explicitly stated in the text?\\

After your analysis, provide your final answer as either 'Yes' (there is cross-modality coreference) or 'No' (there is not) within <answer> tags. Remember, you're looking specifically for references to people in a photo, not objects or other entities.
\end{examplebox}

\section{Human Annotation Details}\label{appen:anno}
\subsection{Annotation Guidance}

\begin{itemize}[noitemsep,topsep=1pt]
    \itemsep 0em
    \item \textbf{General Instructions}: Your task is to evaluate the quality of the provided summary from different evaluation aspects. The summary should consider information from both visual and textual modalities of the dialogue. Please carefully read the dialogue and the summary, and answer the questions by considering the following aspects.

    Aspects
    \begin{enumerate}[noitemsep,topsep=1pt]
        \itemsep 0em
        \item Summary Coherence
        \item Summary Conciseness
        \item Key information Coverage
        \item Information Balance between two modalities
        \item Summary Topic Progression
        \item Summary Faithfulness
    \end{enumerate}

    \item \textbf{Summary Coherence}: Summary coherence in an image-sharing dialogue refers to how effectively the summary integrates and organizes information from both the visual elements of the image and the textual content of the dialogue. A coherent summary should seamlessly combine visual and textual information, presenting it in a logically connected and natural way. In contrast, an incoherent summary will appear fragmented or disjointed, failing to establish clear relationships between the modalities. The coherence rating uses a scale from 1 to 5, where 1 indicates low coherence and 5 indicates high coherence.

    Important: This Coherence here goes beyond typical textual coherence, which primarily focuses on sentence transitions and flow. Instead, it emphasizes the structured and meaningful integration of information across both visual and textual modalities.

    \begin{enumerate}[noitemsep,topsep=1pt]
        \itemsep 0em
        \item \textbf{Score 1, Poor Coherence}: The summary lacks any meaningful integration between the dialogue and image content, presenting them as entirely separate, disjointed or unrelated. The summary may seem confusing, with no logical flow or cohesive narrative.
        \item \textbf{Score 2, Limited Coherence}: The dialogue and image descriptions appear mostly disjointed or unconnected. There are significant gaps or abrupt transitions that make the summary difficult to follow as a unified piece.
        \item \textbf{Score 3, Moderate Coherence}: The integration is only partial. Some aspects of the summary feel disconnected. The connection between image and dialogue feels somewhat forced or inconsistent, with occasional abrupt transitions.
        \item \textbf{Score 4, Good Coherence}: The summary effectively integrates most of the key details from the dialogue and the image. There is a strong overall transition between the image descriptions and dialogue content, though there may be minor parts that feel less cohesive.
        \item \textbf{Score 5, Excellent Coherence}: The summary fully integrates information from both the dialogue and the image in a seamless, natural way. In the summary, there are no disjointed sections or unnatural transitions.
    \end{enumerate}

    \textbf{Summary Examples}:
    
    Excellent Coherence (score 5): "Speaker 0 describes volunteering at a homeless shelter and shares a photo of a stew they made for the residents. While the stew may have looked unappetizing (as seen in the photo), Speaker 0 assures Speaker 1 that it tasted delicious and that at least one person ate some. The conversation highlights Speaker 0's positive attitude towards volunteering, despite the challenges, and the belief that even small acts of kindness are worthwhile."

    Limited Coherence (score 2): "In the dialogue, Speaker 0 shares their experience volunteering at a homeless shelter, mentioning the fulfilling nature of the work despite initial nervousness. They also share a humorous anecdote about a stew they made, which looked unappealing but was reportedly delicious. Speaker 1 expresses admiration for Speaker 0\'s efforts and agrees that food doesn\'t have to look good to taste good. The conversation concludes with Speaker 0 returning to their volunteer work, and Speaker 1 wishing them a good night. The shared photo shows the "glorious" looking stew, emphasizing the contrast between its appearance and taste ."
    
    \item \textbf{Summary Conciseness}: Summary conciseness evaluates how efficiently a summary conveys the essential information without unnecessary verbosity or redundancy. A concise summary is clear, brief, and avoids repeating ideas, using straightforward sentence structures to focus only on the key points. In contrast, a verbose summary includes redundant information, overly complex phrasing, and irrelevant details, which obscure the main ideas.

     Important: When evaluating conciseness, focus on word choice, phrasing, and the elimination of unnecessary details. Do not penalize the summary for missing key information; this will be assessed separately under the Key Information Coverage aspect.

    \begin{enumerate}[noitemsep,topsep=1pt]
        \itemsep 0em
        \item \textbf{Score 1, Extremely Verbose}: The summary is excessively wordy, with frequent repetition and unnecessary details that make it difficult to focus on the main points.

        \item \textbf{Score 2, Somewhat Verbose}: The summary contains excessive words, complex sentence structures, or repeated information. The lack of efficiency reduces its clarity and readability.

        \item \textbf{Score 3, Somewhat Concise}: The summary has occasional redundancy or elaboration, but these do not significantly impact its overall readability or focus.

        \item \textbf{Score 4, Good Conciseness}: The summary is generally efficient and clear, with minimal repetition or extra words. Minor improvements could make it even more concise.

        \item \textbf{Score 5, Excellent Conciseness}: The summary is perfectly concise, with every word serving a purpose. It conveys all key information directly and efficiently, without any superfluous content.
    \end{enumerate}

    \textbf{Summary Examples}:
    
    Excellent Conciseness (score 5): "Speaker 0 described their fulfilling volunteer experience at a homeless shelter, sharing a photo of the unappealing-looking stew they made. Despite its appearance, the stew was delicious, and at least one person sampled it. The conversation highlighted the rewarding aspects of volunteering, even when facing challenges, and the importance of personal effort."
    
    Somewhat Verbose (score 2): "In this heartwarming conversation, we see a dialogue between two individuals discussing volunteer work at a homeless shelter. The volunteer shares their experience, expressing fulfillment despite initial nervousness about helping those in need. They emphasize the importance of doing one's best and not feeling guilty. The conversation takes an amusing turn when the volunteer mentions their less-than-appetizing-looking stew, which they claim tasted delicious despite its appearance. The shared photo in the dialogue shows a man serving the stew, which indeed looks quite unappetizing with its brown, mushy appearance. Despite its unappealing look, the volunteer insists it tasted good and managed to convince at least one person to eat it. The conversation ends on a positive note, with the volunteer returning to their work at the shelter."

    \item \textbf{Summary Critical Information Coverage}: Critical Information Coverage evaluates how effectively the summary captures the essential breadth and depth of key information from the source material. Key information includes details that either describe the core event or significantly contribute to the main ideas of the dialogue. The coverage is assessed across three sub-aspects:
    \begin{enumerate}[noitemsep,topsep=1pt]
        \itemsep 0em
        \item \textbf{Visual Critical Information Coverage}: How well the summary incorporates key details from the visual elements.

        \item \textbf{Textual Critical Information Coverage}: How well the summary captures key points from the textual dialogue.

        \item \textbf{Overall Critical Information Coverage}: The extent to which the summary integrates key information from both modalities.
    \end{enumerate}

    Scoring Criteria: 

    \begin{enumerate}[noitemsep,topsep=1pt]
        \itemsep 0em
        \item \textbf{Score 1: Poor Coverage}: Only $0-20\%$ of the critical information is included in the summary.

        \item \textbf{Score 2: Limited Coverage}: Only $20-40\%$ of the critical information is covered, leaving significant gaps.

        \item \textbf{Score 3: Moderate Coverage}: About $40-60\%$ of the critical information is included, with noticeable omissions.

        \item \textbf{Score 4: Good Coverage}: Approximately $60-80\%$ of the critical information is captured, with minor gaps.

        \item \textbf{Score 5: Excellent Coverage}: Between $80-100\%$ of the critical information is effectively captured, with minimal or no significant omissions.
    \end{enumerate}

    \item \textbf{Summary Modality Information Balance}: Modality Information balancing measures how well a system balances information derived from different modalities, e.g. image and textual dialog. A well-balanced response appropriately leverages visual content and textual context without over-focusing on either source.
    
    Important: When assessing modality balance, note that achieving a 50:50 ratio between modalities is not the goal. The ideal balance is dynamic and depends on the context of the dialogue. A perfectly balanced summary appropriately reflects the critical information from both modalities, based on their importance in the given context.

    The Information Balancing rating uses a bipolar scale from 1 to 7, where:
    
    \begin{enumerate}[noitemsep,topsep=1pt]
        \itemsep 0em
        \item \textbf{Score 1, Extremely textual dialog heavy}: The summary entirely ignores the visual content, resembling pure textual summarization.

        \item \textbf{Score 2, Moderately textual dialog heavy}: The summary predominantly focuses on textual dialogue, giving limited attention to the visual content.

        \item \textbf{Score 3, Slightly textual dialog heavy}: The summary is somewhat skewed toward textual information but includes some visual context.

        \item \textbf{Score 4, Good Balance}: The summary achieves a well-balanced integration of visual and textual information, capturing all critical details from both modalities without unnecessary emphasis on either.

        \item \textbf{Score 5, Slightly image heavy}: The summary leans slightly toward visual content, with textual details less emphasized.

        \item \textbf{Score 6, Moderately image heavy}: The summary predominantly focuses on visual information, giving limited attention to textual dialogue.

        \item \textbf{Score 7, Extremely image heavy}: The summary exclusively discusses visual content, resembling image captioning rather than a balanced summary.
    \end{enumerate}

    \item \textbf{Summary Topic Progression}: Topic progression evaluates how well the summary captures the flow of topics discussed in a dialogue. Summary with good topic progression should ensure smooth transitions between topics and accurate attribution of shared images to the appropriate parts of the dialogue.

    \begin{enumerate}[noitemsep,topsep=1pt]
        \itemsep 0em
        \item \textbf{Score 1, Poor Topic Progression}: The summary fails to capture the dialogue’s topic flow, presenting an unordered or incoherent sequence. Shared images are ignored or entirely misattributed.

        \item \textbf{Score 2, Limited Topic Progression}: The summary reflects some topics but skips key transitions, leading to a disjointed or wrongly-ordered depiction of the dialogue. Shared images are often inaccurately attributed.

        \item \textbf{Score 3, Moderate Topic Progression}: The summary reflects the main topics but misses or simplifies some transitions, resulting in a less coherent depiction of the topic flow. Image attribution may be partially inaccurate.

        \item \textbf{Score 4, Good Topic Progression}: The summary captures the overall flow of topics and most transitions accurately. Shared images are generally well-attributed, with only minor inconsistencies.

        \item \textbf{Score 5, Excellent Topic Progression}: The summary accurately captures the flow of topics in the dialogue, including all key transitions and the logical sequence of ideas. Shared images are attributed accurately to their corresponding topics.
    \end{enumerate}

    \textbf{Summary Example}:
    
    Good Topic Progression (score 4): "Two friends discuss the speaker's volunteer experience at a homeless shelter. The speaker shares their positive experience, emphasizing that their best efforts are always enough. They mention making a stew that looked unappetizing but tasted good, sharing a photo of the dish. The friend expresses admiration for the volunteer work and agrees that appearances can be deceiving when it comes to food. The conversation ends with the speaker returning to their volunteer duties."

    Limited Topic Progression (score 2): "Speaker 0, volunteering at a homeless shelter, chats with Speaker 1 about their experience. While Speaker 0 initially expresses some self-doubt about the less-than-appealing appearance of the stew they made (a shared photo shows a steaming pot of stew and a plate of it being served, highlighting its unassuming look) , they ultimately feel fulfilled by their efforts and emphasize that doing their best is enough. Speaker 1 affirms this sentiment, pointing out that food can look unappetizing yet taste delicious. The conversation concludes with Speaker 0 returning to their volunteer work."

\end{itemize}

\subsection{Annotation Interface}
\begin{figure*}[htbp]
  \centering
  \includegraphics[page=1, width=\textwidth]{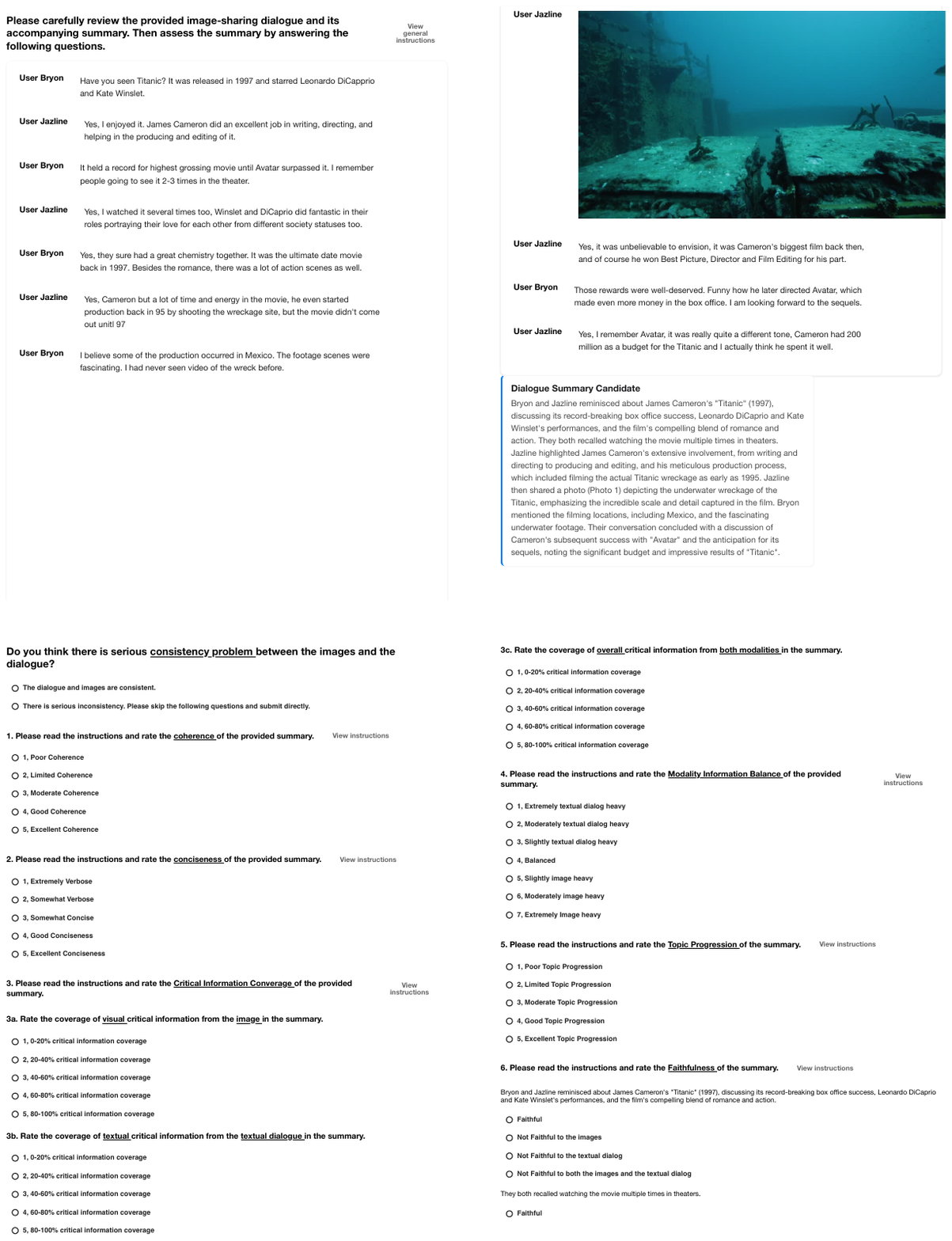}
  \caption{Annotation interface example – Page 1}
  \label{fig:interface_1}
\end{figure*}

\begin{figure*}[htbp]
  \centering
  \includegraphics[page=2, width=\textwidth]{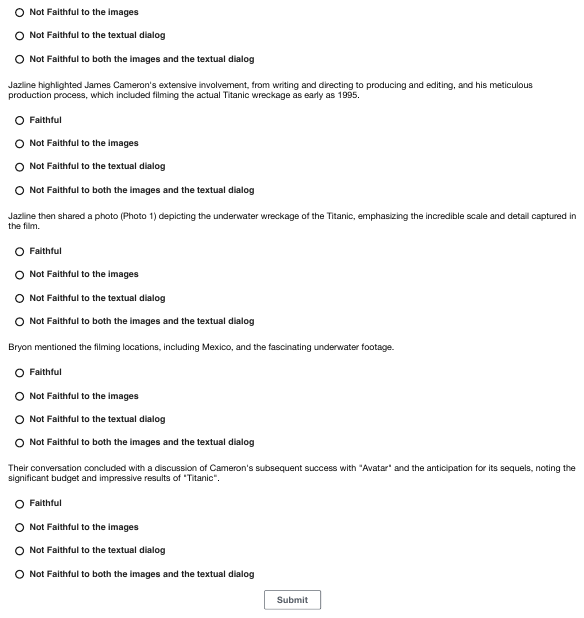}
  \caption{Annotation interface example – Page 2}
  \label{fig:interface_2}
\end{figure*}

We provide an example of our human annotation interfaces in Fig.~\ref{fig:interface_1} and Fig.~\ref{fig:interface_2}.

\subsection{Human Annotation Bypass Mechanism}\label{appen:anno-human}
For each annotation task, annotators are first asked to assess whether the image and the accompanying dialogue are consistent, as presented in the annotation interface. If an annotator determines that the image and dialogue are not naturally paired, they are allowed to skip the remaining annotation steps for that item. Each dialogue is evaluated by three independent annotators. If two or more annotators judge the dialogue to be inconsistent with the image, the dialogue is excluded from the dataset. As a result, the final dataset comprises $93.1\%$ dialogues with three annotations and $6.9\%$ with only two annotations.

\section{Additional Faithfulness Annotation}\label{appen:faith_anno}
In this section, we describe the process used to conduct an additional round of human annotations to resolve cases where no majority vote was reached in the initial annotation round. Specifically, there are 62 summary sentences for which no majority agreement was obtained.

During the original annotation phase, each sentence was evaluated by three annotators. However, since our faithfulness annotation scheme includes four possible label categories (as detailed in Appendix~\S\ref{appen:anno}), it is possible for the votes to be evenly distributed or split in such a way that no single label received a majority.

To resolve these cases, we engaged two additional professional annotators. These annotators independently reviewed all 62 summary sentences lacking a majority vote. When both annotators agreed on a label, their consensus was adopted as the final annotation for that sentence. This approach ensures consistency and reliability in the dataset while maintaining the integrity of the original annotation framework.

\section{Summary-Level Faithfulness Annotation Aggregation}\label{appen:faith_aggre}

To derive summary-level faithfulness annotations from sentence-level labels, we apply the following aggregation rules:

\begin{itemize}[noitemsep,topsep=1pt] 
    \itemsep 0em 
    \item A summary is labeled as faithful only if all of its constituent sentences are individually annotated as faithful. 
    \item A summary is labeled as not faithful to both modalities if any of its sentences is annotated as such, or if the summary contains at least one sentence labeled not faithful to image and at least one labeled not faithful to text. 
    \item If a summary contains only sentence-level annotations of a single unfaithful type—either not faithful to image or not faithful to text—and no sentences are labeled as unfaithful to both, then the summary inherits that specific unfaithful label accordingly. 
\end{itemize}

These rules ensure that the summary-level annotation accurately reflects the integrity of its individual components across visual and textual modalities.

\end{document}

%% file: tables/table_pointwise.tex
\begin{table*}[h]
\centering
\setlength{\extrarowheight}{2pt}
\renewcommand{\arraystretch}{0.85}
\setlength{\tabcolsep}{4pt}
\resizebox{\linewidth}{!}{
    \begin{tabular}{lcccccccccccccc}
    \hline
     \multirow{2}{*}{Models} & \multicolumn{2}{c}{COH} & \multicolumn{2}{c}{CON} & \multicolumn{2}{c}{COV-I} & \multicolumn{2}{c}{COV-T} & \multicolumn{2}{c}{COV-O} & \multicolumn{2}{c}{BAL} & \multicolumn{2}{c}{PROG} \\
     & $\rho \uparrow$ & MSE $\downarrow$ & $\rho \uparrow$ & MSE $\downarrow$ & $\rho \uparrow$ & MSE $\downarrow$ & $\rho \uparrow$ & MSE $\downarrow$ & $\rho \uparrow$ & MSE $\downarrow$ & $\rho \uparrow$ & MSE $\downarrow$ & $\rho \uparrow$ & MSE $\downarrow$ \\
    \cmidrule(lr){1-1} \cmidrule(lr){2-3} \cmidrule(lr){4-5} \cmidrule(lr){6-7} \cmidrule(lr){8-9} \cmidrule(lr){10-11} \cmidrule(lr){12-13} \cmidrule(lr){14-15}
\rowcolor{gray!20} Fine-tuned model &&&&&&&&&&&&&& \\ 
LLaVa-Critic & -5.2 & 0.56 & 3.7 & 0.65 & 7.0 & 0.90 & -1.5 & 0.69 & 1.9 & 0.59 & 4.2 & 0.64 & 4.9 & 0.38 \\ \hline 
\rowcolor{gray!20} GPT-4o-mini &&&&&&&&&&&&&& \\ 
MLLM-as-Judge & 9.1 & 0.41 & 5.6 & 0.57 & 12.0 & 0.96 & -0.4 & 0.50 & -1.7 & 0.58 & 22.2 & 1.47 & 8.0 & 0.60 \\ 
Image-prompt & 1.6 & 0.58 & 6.2 & 0.66 & 12.0 & 2.59 & 4.1 & 0.83 & 4.8 & 1.17 & 8.7 & 1.59 & 1.2 & 0.63 \\ 
Checklist-CoT & 4.4 & 0.69 & -1.7 & 0.91 & 19.6 & 1.19 & 5.3 & 0.67 & 7.2 & 0.87 & 14.7 & 1.62 & 1.6 & 0.90 \\ 
\hline 
\rowcolor{gray!20} Gemini-1.5-flash &&&&&&&&&&&&&& \\ 
MLLM-as-Judge & -3.0 & 0.65 & -0.8 & 0.59 & 3.4 & 0.84 & -3.7 & 0.59 & 5.0 & 0.73 & 5.5 & 0.65 & -2.6 & 0.63 \\ 
Image-prompt & -6.6 & 0.64 & 4.0 & 0.57 & 20.1 & 2.62 & 3.7 & 0.99 & 1.6 & 1.31 & -2.1 & 0.80 & -1.9 & 0.69 \\ 
Checklist-CoT & -4.8 & 0.91 & 5.5 & 1.10 & 7.4 & 0.97 & -0.9 & 0.74 & 4.3 & 0.70 & 1.4 & 0.67 & 1.8 & 0.58 \\ 
\hline 
\rowcolor{gray!20} Qwen-vl-max &&&&&&&&&&&&&& \\ 
MLLM-as-Judge & -2.8 & 0.29 & 4.5 & 0.56 & 1.8 & 0.91 & -2.3 & 0.67 & 2.5 & 0.61 & 1.8 & 0.64 & 3.6 & 0.30 \\ 
Image-prompt & 1.7 & 0.36 & -1.0 & 0.56 & 4.7 & 1.49 & -3.0 & 0.70 & -3.2 & 0.70 & 2.6 & 0.76 & 4.4 & 0.32 \\ 
Checklist-Eval & 2.8 & 0.45 & -5.1 & 0.58 & 5.8 & 0.70 & 6.2 & 0.59 & -2.7 & 0.55 & -2.0 & 0.65 & 3.6 & 0.40 \\ 
\hline
    \end{tabular}
}
\caption{Score-based evaluation results for MLLM evaluation methods. We report Spearman correlations ($\rho$) and Mean Squared Error (MSE) across seven evaluation aspects. The results indicate that state-of-the-art multimodal evaluation methods show limited alignment with human judgments when assessing high-quality summaries.
}
\label{tab:pointwise}
\end{table*}

%% file: tables/table_pairwise.tex
\begin{table*}[h]
\centering
\begin{minipage}{0.59\textwidth}
    \setlength{\extrarowheight}{2pt}
    \renewcommand{\arraystretch}{0.85}
    \setlength{\tabcolsep}{4pt}
    \resizebox{\linewidth}{!}{
        \begin{tabular}{lccccccc}
            \hline
             Models & COH & CON & COV-I & COV-T & COV-O & BAL & PROG \\
            \hline
            \rowcolor{gray!20} GPT-4o-mini &&&&&&& \\ 
            MLLM-as-Judge & 50.6 & 61.8 & 63.4 & 54.7 & 57.8 & 53.5 & 50.8 \\ 
            Image-prompt & 49.8 & 61.2 & 62.8 & 54.9 & 59.0 & 50.5 & 47.8 \\ 
            Checklist-CoT & 48.9 & 62.4 & 63.6 & 52.9 & 55.8 & 53.9 & 48.8 \\ 
            \hline 
            \rowcolor{gray!20} Gemini-1.5-flash &&&&&&& \\ 
            MLLM-as-Judge & 47.7 & 58.6 & 61.1 & 47.6 & 51.0 & 57.4 & 45.8 \\ 
            Image-prompt & 47.9 & 59.8 & 56.0 & 46.6 & 47.1 & 50.7 & 47.8 \\ 
            Checklist-CoT & 47.5 & 52.7 & 60.8 & 46.8 & 49.1 & 55.8 & 46.2 \\ 
            \hline 
            \rowcolor{gray!20} Qwen-vl-max &&&&&&& \\ 
            MLLM-as-Judge & 51.6 & 56.3 & 59.8 & 55.1 & 55.0 & 49.2 & 51.3 \\ 
            Image-prompt & 51.0 & 53.2 & 56.0 & 55.0 & 54.4 & 48.4 & 50.8 \\ 
            Checklist-CoT & 51.3 & 54.3 & 59.5 & 60.2 & 57.8 & 52.1 & 49.3 \\ 
            \hline
        \end{tabular}
    }
    \caption{Pairwise comparison evaluation results for MLLM evaluation methods. We report accuracy across seven evaluation aspects. The results indicate that SOTA multimodal evaluation methods generally struggle to perform pairwise comparisons that align with human preferences. While performance on conciseness and coverage is slightly better than other aspects, the improvement is only marginal.}
    \label{tab:pairwise}
\end{minipage}%
\hfill
\begin{minipage}{0.38\textwidth}
    \vspace{-2mm}
    \resizebox{\linewidth}{!}{%
        \begin{tabular}{lcccc}
            \hline
             \multirow{2}{*}{Models} & \multicolumn{2}{c}{Summ.-lvl} & \multicolumn{2}{c}{Sent.-lvl} \\
             & BAcc.& F1 & BAcc.& F1  \\
             \cmidrule(lr){1-1} \cmidrule(lr){2-3} \cmidrule(lr){4-5}
            \rowcolor{gray!20} Fine-tuned model &&&& \\ 
            LLaVa-Critic & 26.6 & 25.1 & 27.9 & 24.4 \\ 
            \hline 
            \rowcolor{gray!20} GPT-4o-mini &&&& \\ 
            MLLM-as-Judge & 27.3 & 22.3 & 15.5 & 16.5 \\ 
            Image-prompt & 29.8 & 26.6 & 18.7 & 11.9 \\ 
            Checklist-CoT & 29.5 & 10.7 & 22.8 & 2.1 \\ 
            \hline 
            \rowcolor{gray!20} Gemini-1.5-flash &&&& \\ 
            MLLM-as-Judge & 22.2 & 21.9 & 13.3 & 23.3 \\ 
            Image-prompt & 27.5 & 23.8 & 21.9 & 20.0 \\ 
            Checklist-CoT & 24.2 & 22.7 & 22.2 & 15.7 \\ 
            \hline 
            \rowcolor{gray!20} Qwen-vl-max &&&& \\ 
            MLLM-as-Judge & 24.0 & 22.5 & 21.5 & 24.3 \\ 
            Image-prompt & 24.5 & 22.5 & 22.8 & 24.4 \\ 
            Checklist-CoT & 27.3 & 21.6 & 17.5 & 20.4 \\ 
            \hline
        \end{tabular}
    }
    \caption{Multimodal faithfulness evaluation results with Balanced Accuracy (BAcc.) and F1 score at both the summary and sentence levels.}
    \label{tab:faithfulness}
\end{minipage}%

\end{table*}